%% file: 0+main.tex
\pdfoutput=1

\documentclass[11pt]{article}

\usepackage{naacl2021}
\usepackage[T1]{fontenc}
\usepackage{times}
\usepackage{latexsym} 
\usepackage[utf8]{inputenc} 
\usepackage{soul} 
\usepackage{amsmath} 
\usepackage{booktabs} 
\usepackage{color}
\usepackage{xcolor} 
\usepackage{enumitem}
\usepackage{xspace} 
\usepackage{url}
\usepackage{graphicx}
\usepackage{microtype} 
\usepackage{CJKutf8} 

\newcommand{\comment}[1]{{}\ignorespaces}
\newcommand\num{\mbox{five}\xspace} 
\newcommand\our{\mbox{Sensei}\xspace} 
\newcommand\ourfw{\mbox{\our{}-FW}\xspace} 
\newcommand\ourbw{\mbox{\our{}-BW}\xspace} 
\newcommand\ourgs{\mbox{\our{}-GS}\xspace} 
\newcommand\sdh{\textbf{A}\xspace}
\newcommand\ibm{\textbf{B}\xspace}
\newcommand\apm{\textbf{C}\xspace}
\newcommand\ebu{\textbf{D}\xspace}
\newcommand\sod{\textbf{E}\xspace}

\title{\our: Self-Supervised Sensor Name Segmentation}

\author{
  Jiaman Wu, Dezhi Hong, Rajesh Gupta, Jingbo Shang\\
  Computer Science \& Engineering, University of California, San Diego \\
  \{j4wu, gupta, dehong, jshang\}@eng.ucsd.edu
}

\begin{document}
\maketitle

\begin{abstract}
    \input{0-abs}
\end{abstract}
\input{1-intro}
\input{3-method}
\input{4-eval}
\input{2-related}

\input{5-con}

\section*{Acknowledgement}
This work was supported in part by National Science Foundation 1940291 and 2040727. Any opinions, findings, and conclusions or recommendations expressed herein are those of the authors and should not be interpreted as necessarily representing the views, either expressed or implied, of the U.S. Government. The U.S. Government is authorized to reproduce and distribute reprints for government purposes notwithstanding any copyright annotation hereon.





\bibliography{reference}
\bibliographystyle{acl_natbib}

\end{document}

%% file: 0-abs.tex
A sensor name, typically an alphanumeric string, encodes the key context (e.g., function and location) of a sensor needed for deploying smart building applications.
Sensor names, however, are curated in a building vendor-specific manner using different structures and vocabularies that are often esoteric.
They thus require tremendous manual effort to annotate on a per-building basis; even to just segment these sensor names into meaningful chunks.
In this paper, we propose a \emph{fully automated} self-supervised framework, \our, which can learn to segment sensor names without any human annotation.
Specifically, we employ a neural language model to capture the underlying sensor naming structure and then induce self-supervision based on information from the language model to build the segmentation model.
Extensive experiments on \num real-world buildings comprising thousands of sensors demonstrate the superiority of \our over baseline methods.

%% file: 1-intro.tex
\section{Introduction}

Sensor name segmentation, aiming at partitioning a sensor name string into a few semantic segments, is an essential task to enable smart building technologies~\cite{weng2012buildings}, as these technologies fundamentally rely on understanding the context of sensory data.
For example, 
to increase the airflow in a room in view of the ongoing COVID-19 pandemic, one needs to locate the airflow control point of the room. 
To obtain such context, one needs first to be able to understand the sensor names, which are encoded as a concatenation of \emph{segments}. 
Thus, correctly segmenting sensor names into meaningful chunks is a key first step towards such understandings; 

\begin{figure}[t]
	\centering
	\includegraphics[width=0.95\linewidth]{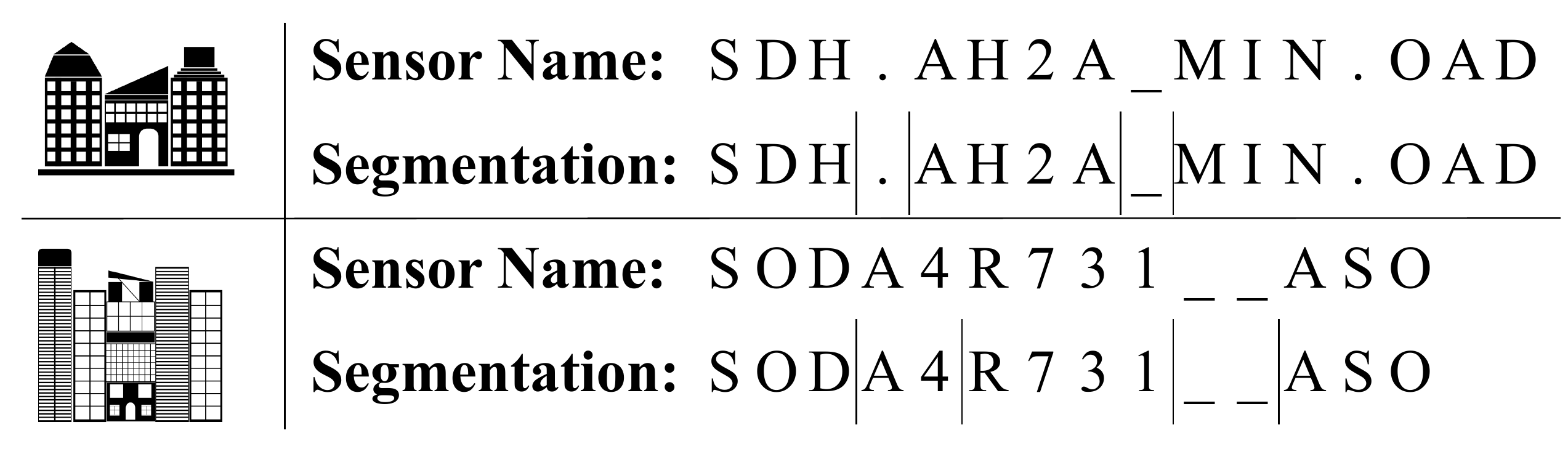}
    \caption{
    Example sensor names in two buildings and their segmentation results.
    Sensor names adopt distinctive structures and vocabularies in different buildings, thus requiring manual effort to interpret. 
    }
    \label{fig:sensorNameExamples}
\end{figure}

As illustrated in Figure~\ref{fig:sensorNameExamples}, a sensor name is typically a sequence of alphanumeric characters---there are multiple segments, each encoding key context about the sensor (building name, location, sensor type, etc).
For example, the sensor name \texttt{SODA4R731\_\_ASO} should be segmented as \texttt{SOD} (building name), \texttt{A4} (equipment id), \texttt{R731} (room id), and \texttt{ASO} (measurement type -- area temperature setpoint). 
Note that the meanings of the same punctuation 
may vary;
for example, `\_' can be a delimiter or part of a segment.

\begin{figure*}[t]
    \centering
    \includegraphics[width=.98\linewidth]{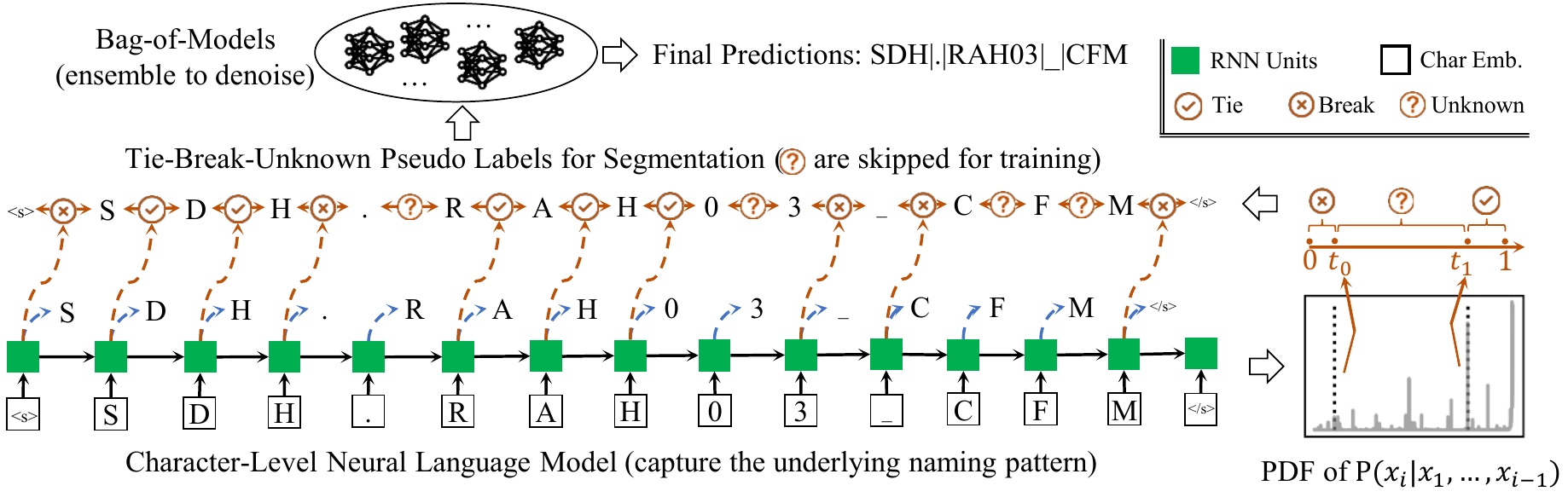}
    \caption{
    Overview of \our.
    We induce pseudo labels for segmentation using the transition probabilities from a character-level neural language model.
    The hidden states from the language model are also used when training the segmentation model.
    }
\label{fig:overview}
\end{figure*}

Currently, sensor name segmentation requires domain knowledge and tedious manual effort due to its building-specific nature. 
Sensor names are created by building vendors, and as we see from Figure~\ref{fig:sensorNameExamples}, in different buildings they usually adopt distinctive structures and vocabularies that are often esoteric.
Typically, to build a sensor name segmentation model, it involves a technician with the domain expertise to comprehend these sensor names and then design rules to segment and annotate these names; no universal pre-defined parsing rules such as regular expressions exist for sensor names.
Therefore, it remains a major obstacle to the wide adoption of smart building technologies from both cost and efficiency perspectives~\cite{bhattacharya2015short}. 

We need an automated solution for sensor name segmentation.
Despite the recent progress in applying active learning~\cite{schumann2014towards,hong2015clustering,balaji2015zodiac,koh2018scrabble,shi2019evaluation} and transfer learning~\cite{hong2015building,jiao2020senser} to sensor name interpretation, all these methods still require human annotation effort and thus they are not \emph{fully} automated.

In this paper, given all the sensor names in a building, we propose a novel self-supervised segmentation framework, \our, to segment these names into meaningful chunks \emph{without any human effort}.
Doing so would facilitate the process of understanding sensor context and make it fundamentally scalable. 
Figure~\ref{fig:overview} presents an overview.

We draw inspiration from a key observation that when creating the sensor names within one building, technicians would follow some underlying naming patterns. For instance, in some buildings, the sensor name often starts with the building name, followed by the room id and type of measurement.
Also, technicians would use similar phrases to express the same concept (e.g., ``temperature'' would be encoded as ``T'', ``temp'', or ``ART''), at least within the same building.

Based on this observation, in \our, we first employ a character-level neural language model~\cite{karpathy2015visualizing} to capture the latent generative pattern in sensor names.
This language model learns the probability of observing a character in the sensor name given all the preceding characters.
Intuitively, the segment boundaries in a sensor name should highly correlate with this probability.
Frequent transitions would have a higher probability than the infrequent ones, which might well imply the start of another segment.
Therefore, we induce pseudo segmentation labels by setting a pair of thresholds on these transition probabilities, and then build a binary classifier to segment sensor names upon their contextualized representations produced by the language model.
Since these pseudo labels may contain noise, we create an ensemble of independent classifiers, each trained on a uniformly random subset of the pseudo labels, in order to further improve the efficacy.

To the best of our knowledge, \our is the first framework for sensor name segmentation without human annotation.
We conduct extensive experiments on \num different buildings with thousands of sensors. 
Our main contributions are as follows:
\begin{itemize}[leftmargin=*,nosep]
    \item We study an important problem of \emph{fully automated} sensor name segmentation.
    \item We propose a novel self-supervised framework \our, which leverages a neural language model to capture the underlying naming patterns in sensor names and produces pseudo segmentation labels for training binary classifiers.
    \item We conduct extensive experiments on \num real-world buildings comprising thousands of sensor names. 
    \our on average achieves about $82\%$ in F$_1$, roughly a 49-point improvement over the best compared method. \comment{?perf}
\end{itemize}
\textbf{Reproducibility}. Our code and datasets are readily available on Github: \url{https://github.com/work4cs/sensei}.

%% file: 3-method.tex
\section{The \our Framework}

Our framework \our consists of three steps:
\begin{itemize}[leftmargin=*,nosep]
    \item Train a neural language model (NLM) at the character level to capture the underlying naming patterns in sensor names;
    \item Generate \texttt{Tie-Break-Unknown} pseudo labels using two thresholds, $t_0$ and $t_1$, decided by inspecting the distribution of transition probabilities (i.e., likelihood of observing the current character given the previous ones);
    \item Train a set of segmentation models based on the pseudo labels to mitigate the effect of noise in these labels.
\end{itemize}
We next elaborate on each step.

\subsection{Language Model for Underlying Patterns}
    
    As sensor names are created by humans (e.g., a technician with knowledge about building particulars), they often follow a certain naming convention (e.g., start with the building name, then room id, and then type).
    In addition, within a building, segments of sensor names corresponding to the same kind of information (e.g., location or function) would use 
    similar phrases; e.g., the concept of ``room'' would be encoded as ``\texttt{RM}'', ``\texttt{R}'', or similar variants.
    A natural solution follows here: we would want to model the generative patterns in these names such that given the characters seen by far we can predict the next one.
    This coincides with the language modeling task in NLP.
    
    Since the sensor name segmentation task works on characters, we adopt a popular character-level neural language model to capture the underlying sensor naming pattern.
    Specifically, we choose the classical Char-RNN~\cite{karpathy2015visualizing} architecture in our design and use LSTM~\cite{hochreiter1997long} as the RNN model.
    Note that, our method is compatible with any character-level neural language models.
    
    Given a character sequence of length $N$, $X = \langle x_1, x_2, \ldots, x_N \rangle$, the Char-RNN learns the probability of observing a character given all the previous characters, namely, $p(x_{i+1} | x_1, x_2, \ldots, x_{i})$.
    During this process, we will obtain an embedding vector $\mathbf{x}_i$ for each character $x_i$, and a hidden state vector $\mathbf{h}_i$ after observing the characters from $x_1$ to $x_i$.
    A softmax layer is then applied to $\mathbf{h}_i$ to predict a distribution $\mathbf{\hat{p}_i}$ over the entire vocabulary:
    \begin{equation*}
        \hat{\mathbf{p_i}}(c) = p(c | x_1, x_2, \ldots, x_{i}) = \frac{\exp\left({\mathbf{w}_{c}^\top \mathbf{h}_i}\right)}{\sum_{c'} \exp\left({\mathbf{w}_{c'}^\top \mathbf{h}_i}\right)},
    \end{equation*}
    where $\mathbf{w}_{c}$ is the linear transformation for character $c$.
    The cross-entropy between $\mathbf{\hat{p}_i}$ and the one-hot encoding of $x_{i+1}$ is used as the loss function for this character.
    
    Given a building, we train the Char-RNN on all its sensor names.
    As each sensor name is independent of each other, we can have the same initial hidden state for each sensor name to ensure sensor names do not interfere with each other. 
    Once the model converges, we apply it to all the sensor names to obtain the character transition probabilities, i.e., $\hat{\mathbf{p_i}}(x_{i+1})$. 
    The perplexity of the trained Char-RNN in our experiments is typically small (i.e., $<0.3$ per batch with batch size 32).
    Therefore, we believe it captures the underlying naming pattern within the input building well.

    \begin{figure}[t]
        \centering
    	\includegraphics[width=0.85\linewidth]{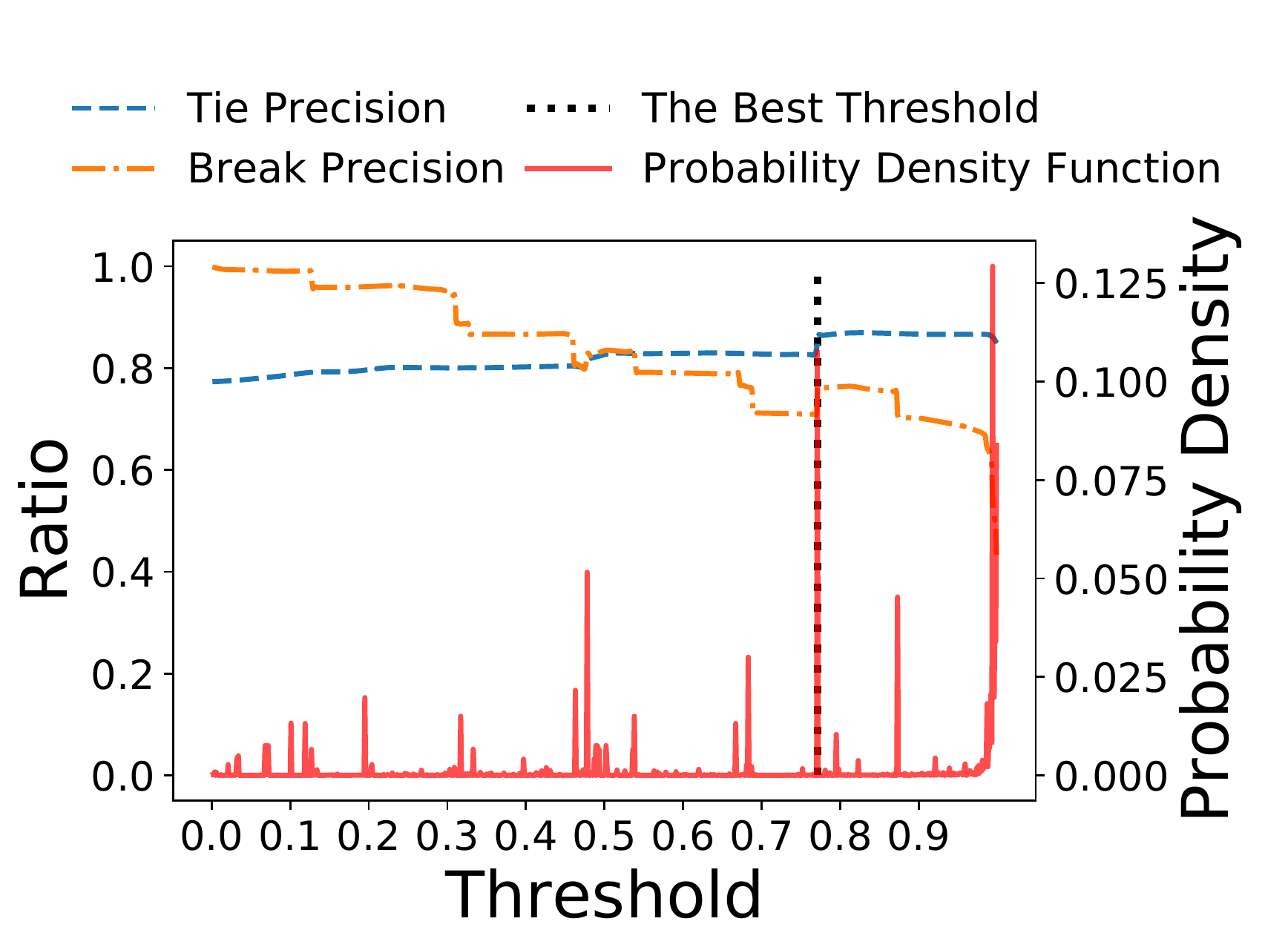}
        \vspace{-3mm}
        \caption{Plots of $\mathbf{\hat{p}_i}(x_{i+1})$ histogram (grey bars) and Tie/Break precision curves for an example building. 
        The ``sweet spot'', achieving a great balance between the tie- and break-precision scores, is highly aligned with the peak in the histogram. 
        }
        \vspace{-3mm}
    \label{fig:threshold}
    \end{figure}
    
\begin{table*}[t]
    \centering
    \small
    \caption{Statistics of \num buildings in our experiments. Building names are replaced by letters from A to E for the anonymity. These builds are from three different campuses: Buildings \sdh and \ibm are from the first campus, \apm and \ebu from the second, and \sod from the third. Example sensor names are also listed for reference.}
    \vspace{-3mm}
    \begin{tabular}{ccccl}
        \toprule
        \textbf{Building} & \textbf{\#Sensors} &  \textbf{\#Segments}  & \textbf{\#Characters} & \textbf{Example Sensor Name} \\
        \midrule
        \sdh & $2,551$ & $2 \sim 5$ & $7 \sim 31$ & \texttt{SDH.AH1\_RHC-4:CTL STPT} \\ 
        \ibm & $1,366$ & $2 \sim 3$ & $6 \sim 28$ & \texttt{1F\_FCU10\_11\_13\_23\_COLLAB} \\ 
        \apm & $1,079$ & $1 \sim 7$ & $4 \sim 34$ & \texttt{AP\&M-CRAC-2-MIG-009.COOLING ON-OFF} \\ 
        \ebu & $1,074$ & $2 \sim 3$ & $7 \sim 35$ & \texttt{EBU3B.3RD FLR AVG CLG-PID1} \\ 
        \sod & $1,335$ & $2 \sim 4$ & $14$ & \texttt{SODC3P09DP\_STA} \\ 
        \bottomrule
    \end{tabular}
    \vspace{-3mm}
    \label{tbl:dataset}
\end{table*}

\subsection{Pseudo Labels from Transition Probabilities}
\label{sec:p_label}

    Inspired by~\cite{shang2018learning}, we use \texttt{Tie} and \texttt{Break} to decide the segmentation results.
    The transition between two adjacent characters ($x_i, x_{i+1}$) is labeled as (1) \texttt{Break} when we should segment after character $x_i$, or (2) \texttt{Tie} otherwise, denoting that the two successive characters belong to the same segment.
    
    For a given character sequence $x_1, x_2, \ldots, x_N$,
    we hypothesize that the transition probability $\mathbf{\hat{p}_i}(x_{i+1})$ obtained from Char-RNN is closely related to the \texttt{Tie/Break} relation between $x_i$ and $x_{x+1}$. 
    Intuitively, the Char-RNN model should produce a high likelihood for common transitions in sensor names, e.g., substrings for building name, room, and common sensor types.
    Therefore, when Char-RNN suggests a low transition probability, the transition is very likely to be a \texttt{Break}; otherwise, the possibility of a \texttt{Tie} becomes higher.
    
    We empirically verify our hypothesis via data analysis of an example building as shown in Figure~\ref{fig:threshold}.
    We present the probability density from histogram of $\mathbf{\hat{p}_i}(x_{i+1})$. 
    In addition, based on the ground-truth segmentation results, we plot the Tie and Break precision curves w.r.t. different thresholds.
    The \emph{Tie Precision} refers to the ratio of Tie transitions among all the transitions above a certain threshold, while the \emph{Break Precision} refers to the ratio of Break transitions among all the transitions below a certain threshold.
    One can observe that the ``turning points'' on the break precision curve are highly correlated to the peaks in the histogram.
    
    If one wants to set up a single threshold on $\mathbf{\hat{p}_i}(x_{i+1})$ to classify all transitions into \{\texttt{Tie}, \texttt{Break}\} in an unsupervised manner, the highest peak in the ``confidence'' interval $[0.550, 0.950]$ on the distribution (e.g., $0.771$ in Figure~\ref{fig:threshold}) would be a good choice to achieve a high F$_1$ score.
    We generalize this threshold selection criterion to the other buildings, and as we shall demonstrate in our experiments, such a selection strategy gives results close to grid search that uses ground-truth labels.

    In addition to \texttt{Tie} and \texttt{Break}, we mark those uncertain transitions as \texttt{Unknown}. 
    We need to decide on two thresholds, $t_0$ and $t_1$, and categorize the transitions according to three transition probability intervals, $[0, t_0]$, $(t_0, t_1)$, and $[t_1, 1]$, denoting \texttt{Break}, \texttt{Unknown}, and \texttt{Tie}, respectively, as the \emph{pseudo labels}.
    We wish these pseudo labels would be of high accuracy while having a sufficient amount of labels.
    Based on our observations, the above single threshold criterion satisfies $t_1$. 
    Considering that \texttt{Break}s are considerably fewer than \texttt{Tie}s, we should decide on a \texttt{Break} more carefully. 
    The highest peak in a narrowed high precision interval $[0.050, 0.150]$ would be appropriate (e.g., $0.101$ in Figure~\ref{fig:threshold}).
\subsection{Ensemble to De-noise Pseudo Labels}

    There could exist errors in these automatically induced pseudo labels, so we leverage the idea of ensemble learning to mitigate the effects of these label errors on the final predictions~\cite{breiman1996bagging}.
    Specifically, we independently sample a subset of pseudo labels to train $K$ binary classifiers and then average their predictions.
    In the pseudo labels, the number of \texttt{Tie} transitions is usually much higher than that of \texttt{Break}. 
    To balance the training data, we sample $\epsilon \cdot M$ \texttt{Tie} and \texttt{Break} labels, respectively, from all the pseudo labels, where $M$ is the number of \texttt{Break} transitions and $\epsilon$ is a small coefficient between 0 to 1 for sampling a subset (e.g., $\epsilon=0.1$).
    Such a sampling strategy makes the label errors less likely to affect every binary classifier, so the final prediction becomes more accurate.
    
    All types of binary classifiers could be used to construct the ensemble, and we adopt a multi-layer Perceptron (MLP) as our binary classifier. 
    For the $i$-th transition, we retrieve the hidden state vector $\mathbf{h}_i$ yielded by the Char-RNN and feed it as input to the MLP.
    The final prediction is the average of predictions from the $K$ classifiers.
    As the training data is sampled in a balanced way, we simply use $0.5$ as the threshold to decide on \texttt{Tie} or \texttt{Break}.
    
    

%% file: 4-eval.tex
\section{Experiments}

    We empirically evaluate \our on datasets from real-world buildings and discuss our results as well as findings from some interesting cases.

\subsection{Datasets and Pre-processing}
    
    To evaluate \our, we collect the sensor names from \num office buildings (named A through E) of four different building vendors at three different sites located in different geographic regions. 
    We also collect the character-level ground-truth labels of these names from their building vendors. 
    We adopt the \texttt{BIO} tagging scheme in generating labels, marking the beginning (\texttt{B}), inside (\texttt{I}), and outside (\texttt{O}) of each segment (e.g., for location or function). 
    The details of each building are summarized in Table~\ref{tbl:dataset}. 

    \paragraph{Digits.}
    The digits in sensor names indicate detailed and specific information such as room or equipment identifiers, so preserving the variety in numbers does not help our segmentation task. 
    Conversely, it disturbs the transition probability distribution and thus confuses the model in predicting the next characters -- the model would only need to learn and recognize the transitions from digit to digit, as opposed to the specific values (e.g., ``\texttt{1}'' to ``\texttt{2}'' or ``\texttt{4}'' to ``\texttt{3}''). 
    Therefore, we replace all numerical digits with the same digit ``0''.

    \paragraph{Punctuation and Whitespace.}
    There are symbols such as underscores and whitespace in sensor names, which are inserted by technicians at the time of metadata construction. 
    We leave them \emph{as-is} for our model to learn their meanings
    because the meanings of these characters vary from case to case.
    This is in fact one of the major challenges in this sensor name segmentation problem.
    For example, the sensor name ``\texttt{SODH1\_\_\_\_\_\_L\_L}'' should be segmented as ``\texttt{SOD$|$H1$|$\_\_\_\_\_\_$|$L\_L}'', with the three segments corresponding to its building name, equipment id, and measurement type, respectively. 
    The underscores between ``\texttt{H1}'' and ``\texttt{L\_L}'' are padded to make the sensor name fixed-length, while the underscore inside ``\texttt{L\_L}'' connects two initial letters (i.e., for a Lead-Lag sensor, commonly existing in water pumps). 


\subsection{Evaluation Metrics}
    
    We evaluate the performance of all the considered methods by the F$_1$, precision, and recall scores.
    A segment is represented as a span with the starting and the ending character indices. 
    A predicted segment is correct if and only if there exists an exactly same segment in the ground truth.
    Therefore, we define the precision and recall as follows:
    \begin{equation*} \label{eq:metric}
        \mbox{prec} = \frac{|\mathcal{S}_{GT}| \cap |\mathcal{S}_{Pred}|}{|\mathcal{S}_{Pred}|},
        \mbox{rec} = \frac{|\mathcal{S}_{GT}| \cap |\mathcal{S}_{Pred}|}{|\mathcal{S}_{GT}|},
    \end{equation*}
    where $\mathcal{S}_{GT}$ is the set of ground-truth spans and $\mathcal{S}_{Pred}$ is the predicted set.
    The F$_1$ score is the harmonic mean of precision and recall.
    We report the averaged F$_1$ score of all sensor names, which is relatively unbiased~\cite{opitz2019macro}.
    
    As we mentioned before, there will be some extra delimiters between segments.
    Therefore, during the evaluation, we ignore segments containing only delimiter(s) in both ground truth and predicted segments.
    When calculating the start and end indices for predicted segments, we also skip their prefix and suffix delimiters.
    The same process here applies to the evaluation of all methods.

\begin{table*}[t]
    \centering
    \caption{Performance of \our and compared methods on the \num test buildings (A-E). }
    \vspace{-3mm}
    \small
        \scalebox{0.97}{
	\setlength{\tabcolsep}{0.9mm}{
    \begin{tabular}{r|ccc|ccc|ccc|ccc|ccc|c}
        \toprule
        \multicolumn{1}{r}{} & \multicolumn{3}{c}{\sdh} & \multicolumn{3}{c}{\ibm} & \multicolumn{3}{c}{\apm}& \multicolumn{3}{c}{\ebu}& \multicolumn{3}{c}{\sod} & \\
        \cmidrule{2-16}
        Methods &  Prec & Rec & F$_1$ & Prec & Rec & F$_1$ &  Prec & Rec & F$_1$ & Prec & Rec & F$_1$ &  Prec & Rec & F$_1$ & Avg F$_1$\\
        \midrule
        Delimiter & 33.21 & 47.18 & 38.47~& 52.61 & 65.87 & 57.80~& 3.10 & 4.44 & 3.56~& 32.00 & 46.60 & 37.73~& 46.54 & 23.95 & 31.51~& \underline{33.81}\\
        NLTK & 18.34 & 31.86 & 22.75~& 0.07 & 0.05 & 0.06~& 3.95 & 4.07 & 3.99~& 20.76 & 27.78 & 23.75~& 0.04 & 0.02 & 0.03 & 10.12\\
        CoreNLP & 17.09 & 13.46 & 14.75~& 0.04 & 0.02 & 0.03~& 39.31 & 30.88 & 34.20~& 15.86 & 10.58 & 12.69~& 0.11 & 0.06 & 0.07~& 12.35\\
        Stanza & 9.30 & 6.95 & 7.82~& 0.0 & 0.0 & 0.0~& 2.51 & 2.75 & 2.57~& 9.21 & 9.09 & 8.92~& 0.0 & 0.0 & 0.0~& 3.86\\
        BayesSeg & 1.74 & 2.17 & 1.92~& 19.72 & 28.54 & 23.25~& 9.72 & 10.18 & 9.88~& 15.05 & 25.16 & 18.82~& 45.07 & 34.16 & 38.84~& 18.54\\
        ToPMine & 16.83 & 31.42 & 21.76~& 27.83 & 38.86 & 31.86~& 14.39 & 30.63 & 19.46~& 2.11 & 4.55 & 2.85~& 15.38 & 26.17 & 19.27~& 19.04\\
        \midrule
        \ourbw & 10.93 & 11.00 & 9.18~& 0.0 & 0.0 & 0.0~& 0.98 & 3.86 & 1.53~& 1.04 & 4.66 & 1.69~& 19.22 & 11.33 & 13.77~& 5.23\\
        \ourfw & 61.17 & 74.45 & 66.56~& 39.97 & 53.40 & 44.84~& 38.58 & 55.58 & 44.81~& 47.94 & 64.65 & 53.78~& 58.38 & 74.18 & 64.87~& 54.97\\
        \ourgs & 61.17 & 74.45 & 66.56~& 79.84 & 80.43 & 79.76~& 38.58 & 55.58 & 44.81~& 47.94 & 64.65 & 53.78~& 58.38 & 74.18 & 64.87~& 61.91\\
        \our & {\bf 87.00} & {\bf 83.64} & \textbf{84.95}~& {\bf 84.81} & {\bf 90.80} & \textbf{86.84}~& {\bf 70.23} & {\bf 77.98} & {\bf 73.21}~& {\bf 78.10} & {\bf 85.77} & {\bf 80.39}~& {\bf 85.81} & {\bf 87.53} & \textbf{86.43}~& \textbf{82.36} \\
        \bottomrule
    \end{tabular}
    }
    }
    \vspace{-2mm}
    \label{tbl:main_perf}
\end{table*}

\subsection{Compared Methods}
    
    We compare \our with the following methods:
    \begin{itemize}[leftmargin=*,nosep]
        \item {\bf Delimiter}. 
            There are punctuation (such as ``-'' and ``\_'') and whitespace characters in sensor names, and they could indicate the boundaries between segments. 
            Therefore, this method segments a sensor name by delimiters (i.e., non-alphanumeric characters). 
            This method mainly serves as a sanity check.
        \item {\bf NLTK TweetTokenizer}. 
            NLTK~\cite{bird2009natural} provides a tweet tokenizer to segment a string into tokens according to predefined regular expressions (regexes). We directly apply it to segment our sensor names. 

        \item {\bf CoreNLP}.
            We adopt the pre-trained tokenizer in the CoreNLP package\footnote{\url{https://stanfordnlp.github.io/CoreNLP/}}~\cite{manning2014stanford}, which adopts the Universal Dependencies\footnote{Universal Dependencies is a framework of annotation guidelines by open community effort. \url{https://universaldependencies.org/}} version 2 (UD v2) standard for segmentation . 
        \item {\bf Stanza}.
            We also adopt Stanza\footnote{\url{https://stanfordnlp.github.io/stanza/}} and use its built-in neural tokenizer~\cite{qi2020stanza} following UD v2.
            This method combines convolutional filters and bidirectional LSTM to realize tokenization and sentence segmentation as a tagging task~\cite{qi-etal-2018-universal}. 
        \item {\bf BayesSeg}.
            Topic segmentation divides a document into topic-coherent segments. 
            An unsupervised Bayesian model, BayesSeg\footnote{\url{https://github.com/jacobeisenstein/bayes-seg}}~\cite{eisenstein2008bayesian}, is used to segment characters of sensor names as a topic segmentation task that decides the boundary between sentences. 
            However, this method requires to manually specify the number of segments, which is a parameter we do not know without human input.
        \item {\bf ToPMine}.
            ToPMine~\cite{el2014scalable} provides a method that groups frequent words into phrases in an unsupervised manner and incorporates these phrases into topic modeling. We adapt the model to work at the character level. That is, we regard each character of sensor names as a word in document and group characters into segments as group words into phrases. 
    \end{itemize}
    Note that, we do not use custom {\bf regular expressions} (regexes) to segment sensor names because they require tremendous manual effort to create in order to exhaustively cover all the possible substring patterns, which deviates from our self-supervised problem setting. Moreover, since different buildings follow different sensor naming conventions, manual effort is required from domain experts to create regexes on a per-building basis, which is a costly process.
    
    \noindent We also compare with two ablations of our method:
    \begin{itemize}[leftmargin=*,nosep]
        \item {\bf \our-Forward (\ourfw)}.
            It leaves out the self-supervised ensemble learning. 
            Specifically, we keep the Char-RNN to obtain the distribution of observing next characters, and then find the \emph{single} threshold as stated in Section~\ref{sec:p_label}.
        \item {\bf \our-Backward (\ourbw)}. 
            This is similar to the forward counterpart. 
            The only difference is that the Char-RNN takes as input the \emph{reversed} sensor names. 
            As we shall see in the results, this method does not add much value to our task due to the intrinsic irregularity of sensor names when examined backward.
    \end{itemize}
    
    \noindent We further examine a method using grid search based on ground truth for threshold tuning to verify the effectiveness of our threshold decision:
    \begin{itemize}[leftmargin=*,nosep]
        \item {\bf \our-GridSearch (\ourgs)}. 
        Compared to \ourfw, this method finds the best threshold for deciding \texttt{Tie} using ground-truth labels, i.e., it searches through all the possible threshold values on the transition probability distribution and picks the one that produces the best segmentation results.
        Note that this method is only used to demonstrate that a single threshold chosen based on the transition distribution (as detailed in Section~\ref{sec:p_label}) gives results reasonably close to the best we can achieve for \our-FW using the ground truth.
    \end{itemize}

\begin{table*}[t]
    \centering
    \caption{Performance of \our using different amounts of sensor names for training. 
    }
    \vspace{-3mm}
    \small
    \begin{tabular}{r|cc|cc|cc|cc|cc}
        \toprule
        \multicolumn{1}{r}{} & \multicolumn{2}{c}{\sdh} & \multicolumn{2}{c}{\ibm} & \multicolumn{2}{c}{\apm}& \multicolumn{2}{c}{\ebu}& \multicolumn{2}{c}{\sod} \\
        \cmidrule{2-11}
        Percentage (\%) &  \#Sensors & F$_1$ & \#Sensors & F$_1$ & \#Sensors & F$_1$ & \#Sensors & F$_1$ & \#Sensors & F$_1$\\
        \midrule
        25 & 637 & 72.67~& 341 & 75.95~& 269 & 39.99~& 268 & 33.50~& 333 & 57.86\\
        50 & 1,275 & 92.28~& 683 & 71.84~& 539 & 48.99~& 537 & 47.77~& 667 & 85.62\\
        75 & 1,913 & 86.38~& 1,024 & 85.04~& 809 & 57.61~& 805 & 70.45~& 1,001 & 85.31\\
        100 & 2,551 & 84.95~& 1,366 & 86.84~& 1,079 & 73.21~& 1,074 & 80.39~& 1,335 & 86.43~\\
        \bottomrule
    \end{tabular}
    \label{tbl:size_perf}
    \vspace{-3mm}
\end{table*}

\subsection{Experimental Setup}

    We modify the Char-RNN library\footnote{\url{https://github.com/sherjilozair/char-rnn-tensorflow}} and use Keras~\cite{chollet2015keras} to implement our method.
    As our method is unsupervised, we do not employ the commonly used early-stopping scheme when training the Char-RNN.
    Instead, we train our models for $100$ epochs and empirically find this to be sufficient. 
    All the thresholds have three decimal places. We assign \texttt{Tie}s as positives and \texttt{Break}s as negatives. For binary classifier, any supervised learning algorithm (e.g., logistic regression, SVM, etc) would accommodate our need in this work.
    We choose a vanilla Multilayer Perceptron with $2$ fully-connected layers, each with $64$ cells. 
    We set the number of binary classifiers in our ensemble, $K$, at $100$.
    The subsampling rate for the ensemble, $\epsilon$, is $10\%$ and for each subsampling, we use pandas with the iteration index as seed.
    Training a \our model on a Colab
    GPU with 12GB RAM takes less than 40 minutes for each building.
    For the other compared methods, we tune at our best based on the recommended settings in their papers or repositories and report the best performance. 

\subsection{Result Analysis}
\label{sec:result_analysis}

    Experimental results for all the methods are summarized in Table~\ref{tbl:main_perf}.
    Overall, \our outperforms all the compared methods significantly, attributed to its strategy of complementing the language model with a self-supervised ensemble classifier.
    Besides the variants of \our, the baseline Delimiter, though simple, has achieved the second best performance among all others methods. 
    On average, Delimiter achieves $33.81\%$ in F$_1$ across all the buildings.
    By contrast, our \our achieves over $80\%$ in F$_1$, which demonstrates a $49$-point improvement over Delimiter.
    
    When looking at the F$_1$ scores of the other baselines, including ToPMine, BayesSeg, and the off-the-shelf tokenizers in NLTK, Stanza, and CoreNLP, they are not competitive; this highlights the need of a solution to our challenging problem. 
    
    The performance of Delimiter also confirms the fact that the semantics of these delimiters are mixed.
    If one recalls the examples in Table~\ref{tbl:dataset}, vendors usually use delimiters in sensor names.
    Sometimes, these delimiters well indicate the segment boundaries.
    However, as we illustrated in the example sensor name ``\texttt{SOD$|$H1$|$\_\_\_\_\_\_$|$L\_L}'', punctuation could be also used within the segment, and therefore simply segmenting at delimiters results in a considerable amount of false positives. 
    
    From \ourfw to \our, there is a significant boost, roughly 27 points in F$_1$ on average. 
    Since the major difference between \our and \ourfw is our self-supervised ensemble learning module, we empirically verified its power.
    
    Comparing \ourfw and \ourbw, one can observe that the forward version performs dramatically better.
    As shown in Table~\ref{tbl:main_perf}, \ourfw performs better than Delimiter, ToPMine, and all the pre-trained tokenizers in all cases. 
    By contrast, \ourbw takes the reversed sensor names as input but performs much worse than \ourfw.
    We notice that this is because there are not sufficient variations in the sensor string patterns when being looked at backward, compared to the forward case.
    For example, there are names like ``\texttt{SODA4R731\_\_ASO}'' and ``\texttt{SODA1R516\_\_VAV}'', and the \ourfw model can see various substrings (e.g., ``\texttt{ASO}'' and ``\texttt{VAV}'') following the common pattern ``\texttt{SODA0R000\_\_}''. 
    Variations as such provide enough information for the model to learn where to segment.
    However, when reversed, the above example becomes ``\texttt{OSA\_\_000R0ADOS}'' and the \emph{prefix} ``\texttt{OSA}'' sees no variations following, which makes it nearly impossible for \ourbw to figure out the right segmentation.
    Consequently, \ourfw better captures generative patterns while \ourbw achieves poor segmentation results.
    
    
    
    Comparing \ourfw and \ourgs, one can observe that, in most cases (4 datasets out of 5), \ourfw finds the best single threshold found by \ourgs. 
    Note that \ourgs utilizes the ground truth to exhaustively search among all the possible thresholds, while \ourfw decides the threshold based on the transition distribution without requiring any labels.
    This small difference in performance indicates that our data-driven threshold finding solution based on the distribution is reasonable and reliable.

    
\subsection{Performance w.r.t. Number of Sensors}
    
    Since our \our framework is fully automated, its performance is solely decided by the amount and variety of available sensor names.
    As shown in Table~\ref{tbl:size_perf}, \our generally gets better performance with more sensor names available with an exception of Building \sdh.
    We hypothesize that the performance relates more closely to the variety of sensor name patterns in the dataset rather than the size.

\subsection{Case Studies and Discussions}
\label{sec:case}
    
    We next showcase some examples that \our correctly segments,
    in order to illustrate its capability.

    
    \paragraph{``Flukes'' for False Positives.} 
    In Building \ibm{}, some of the \texttt{Break}s are recognized as \texttt{Tie}s by \ourfw and \ourgs. For example,
    \begin{center}
         \texttt{0F|\_|SRVC|\_|D0D0D0D00}, \\
         \texttt{GF|\_|SRVC|\_|QR000\_000},
    \end{center}
    are mistakenly segmented as
    \begin{center}
         \texttt{0F\_SRVC|\_|D0D0D0D00}, \\
         \texttt{GF\_SRVC|\_|QR000\_000}.
    \end{center}
    
    By contrast, \our avoids the mistakes by learning the pattern from many other sensor names.
    The following case is a great example. 
    \begin{center}
        \texttt{GF|\_|LGHT|\_|COFFEEDOCK}.
        \texttt{GF|\_|FRONTAISLE|\_|LHS}, \\
        \texttt{0F|\_|FCU\_KWH}. \\
    \end{center}
    There are only $89$ occurrences of ``\texttt{\_|LGHT|}'' compared to $177$ of ``\texttt{\_|SRVC|}''. Thus, with a lower transition probability, it can be recognized as a \texttt{Break} before ``\texttt{\_|LGHT|}''. Many similar cases can teach \our that \texttt{Break} is more likely in this pattern, facilitating its performance.
    
    \paragraph{``Flukes'' for False Negatives.}
    Building \sod contains many cases as follows:
    \begin{center}
         \texttt{SOD|A0|R000|\_\_|ASO}, \\
         \texttt{SOD|A0|R000|\_\_|AGN}.
    \end{center}
    
    \ourfw, and even \ourgs which employs the ground truth, are not able to segment these names correctly; they instead segment them as
    \begin{center}
        \texttt{SOD|A0|R000|\_\_|A|SO}, \\
        \texttt{SOD|A0|R000|\_\_|A|GN},
    \end{center}
    because of the same prefix ``\texttt{SODA0R000\_\_A}''.
    
    By contrast, \our is able to correctly segment them owing to the self-supervised ensemble learning, which is more robust to noise in pseudo labels.
    
    \paragraph{Discussion.}
    We notice that even though \our on average achieves about 80\% in F$_1$, it still has limitations. \our is sensitive to the variation of patterns in datasets---the patterns cannot be too varied or too monotonous.\comment{?perf}

%% file: 2-related.tex
\section{Related Work}

Our work is related to three lines of work, namely, sensor metadata mapping, language model, and phrase mining.

\paragraph{Sensor Metadata Tagging.} 
Sensor Metadata Tagging refers to the process of parsing and annotating the sensor metadata (or sensor name) for understanding a sensor's key context, including the measurement type~\cite{balaji2015zodiac,hong2015clustering}, location~\cite{bhattacharya2015automated}, relationships with others~\cite{koh2018scrabble}, and many more~\cite{schumann2014towards}. The majority body of work exploits an active learning-based procedure~\cite{settles2009active}, where it iteratively selects an ``informative'' and ``representative'' metadata example for a domain expert to label, in order to learn a model to annotate the metadata.
Complementary to the use of textual metadata, there are also efforts exploring the use of time-series data for inferring the sensor context~\cite{koc2014comparison,pritoni2015short}.
While they can significantly reduce the amount of required manual labeling, they still rely on the availability of at least one human annotator to segment, parse, and provide labels.

By contrast, the method proposed in this work is fully automated, i.e., completely removing humans from the process, and we demonstrate its use in an essential first step---segmenting a sensor name string into meaningful substrings.

\paragraph{Language Model and Tokenization.}

Language models originate from the areas of natural language processing and information retrieval~\cite{schutze2008introduction}. 
They aim at modelling the likelihood of observing one token given all the tokens before it, capturing the underlying language patterns.
Recent advances in deep learning have pushed the language modeling from traditional n-gram models to neural language models~\cite{kiros2014multimodal,karpathy2015visualizing,kim2016character,peters2018deep,devlin2018bert}, achieving significantly better performance using recurrent neural networks.

Analogizing sensor names to human languages, we employ neural language models to capture the underlying naming pattern.
As we seek to segment a sensor name string into substrings, we choose the classic Char-RNN model~\cite{karpathy2015visualizing}.
In general, any character-level language models are applicable in our method.

One can also view our problem as tokenization of sensor names.
We thus compare with multiple existing tokenizers provided in NLTK Twitter, Standford CoreNLP~\cite{manning2014stanford}, and Stanza~\cite{qi2020stanza}. 
As we demonstrate in evaluation, our method significantly outperforms these methods in segmenting sensor names.

\paragraph{Phrase Mining.}
Treating characters as words, our problem can be viewed as an unsupervised phrase mining problem with phrasal segmentation as output.
Existing methods mainly leverage statistical signals based on term frequency in the corpus~\cite{deane2005nonparametric,parameswaran2010towards,danilevsky2014automatic,el2014scalable}.  
Among all these methods, ToPMine~\cite{el2014scalable} is arguably the most effective one.
Our method \our significantly outperforms ToPMine in our empirical evaluation.

%% file: 5-con.tex
\section{Conclusions and Future Work}

In this paper, we study the problem of automating building metadata segmentation, which is an important first step to understanding the context of sensor data in buildings; smart building technologies rely on this information.
We present \our, which is a fully automated method without requiring human labels.
\our employs a character-level neural language model to capture the underlying generative patterns in building sensor names. Based on the probability distribution of character transitions (i.e., likelihood of observing the current character give the previous ones), it decides on two thresholds for sifting out examples for which it is confident to be \texttt{Tie} or \texttt{Break}. 
Considering these pseudo-labeled examples as supervision, \our constructs an ensemble of binary classifiers to segment sensor names with the information provided by the language model. 
We conducted experiments on the sensor names from five real-world buildings, and \our on average achieves F$_1$ over $80\%$ in segmenting sensor names, a roughly $49$-point improvement over the best of compared methods.

As future work, collecting a larger collection of sensor metadata to pre-train our language model might significantly improve \our's performance. We also plan to show more usage of \our in standard language tasks in NLP.

%% file: 0+main.bbl
\begin{thebibliography}{33}
\expandafter\ifx\csname natexlab\endcsname\relax\def\natexlab#1{#1}\fi

\bibitem[{Balaji et~al.(2015)Balaji, Verma, Narayanaswamy, and
  Agarwal}]{balaji2015zodiac}
Bharathan Balaji, Chetan Verma, Balakrishnan Narayanaswamy, and Yuvraj Agarwal.
  2015.
\newblock Zodiac: Organizing large deployment of sensors to create reusable
  applications for buildings.
\newblock In \emph{BuildSys}, pages 13--22. ACM.

\bibitem[{Bhattacharya et~al.(2015{\natexlab{a}})Bhattacharya, Ploennigs, and
  Culler}]{bhattacharya2015short}
Arka Bhattacharya, Joern Ploennigs, and David Culler. 2015{\natexlab{a}}.
\newblock Short paper: Analyzing metadata schemas for buildings: The good, the
  bad, and the ugly.
\newblock In \emph{Proceedings of the 2nd ACM International Conference on
  Embedded Systems for Energy-Efficient Built Environments}, pages 33--34. ACM.

\bibitem[{Bhattacharya et~al.(2015{\natexlab{b}})Bhattacharya, Hong, Culler,
  Ortiz, Whitehouse, and Wu}]{bhattacharya2015automated}
Arka~A Bhattacharya, Dezhi Hong, David Culler, Jorge Ortiz, Kamin Whitehouse,
  and Eugene Wu. 2015{\natexlab{b}}.
\newblock Automated metadata construction to support portable building
  applications.
\newblock In \emph{BuildSys}, pages 3--12. ACM.

\bibitem[{Bird et~al.(2009)Bird, Klein, and Loper}]{bird2009natural}
Steven Bird, Ewan Klein, and Edward Loper. 2009.
\newblock \emph{Natural Language Processing with Python}, 1st edition.
\newblock O'Reilly Media, Inc.

\bibitem[{Breiman(1996)}]{breiman1996bagging}
Leo Breiman. 1996.
\newblock Bagging predictors.
\newblock \emph{Machine learning}, 24(2):123--140.

\bibitem[{Chollet et~al.(2015)}]{chollet2015keras}
Fran\c{c}ois Chollet et~al. 2015.
\newblock Keras.
\newblock \url{https://keras.io}.

\bibitem[{Danilevsky et~al.(2014)Danilevsky, Wang, Desai, Ren, Guo, and
  Han}]{danilevsky2014automatic}
Marina Danilevsky, Chi Wang, Nihit Desai, Xiang Ren, Jingyi Guo, and Jiawei
  Han. 2014.
\newblock Automatic construction and ranking of topical keyphrases on
  collections of short documents.
\newblock In \emph{Proceedings of the 2014 SIAM International Conference on
  Data Mining}, pages 398--406. SIAM.

\bibitem[{Deane(2005)}]{deane2005nonparametric}
Paul Deane. 2005.
\newblock A nonparametric method for extraction of candidate phrasal terms.
\newblock In \emph{Proceedings of the 43rd Annual Meeting of the Association
  for Computational Linguistics (ACL’05)}, pages 605--613.

\bibitem[{Devlin et~al.(2018)Devlin, Chang, Lee, and
  Toutanova}]{devlin2018bert}
Jacob Devlin, Ming-Wei Chang, Kenton Lee, and Kristina Toutanova. 2018.
\newblock Bert: Pre-training of deep bidirectional transformers for language
  understanding.
\newblock \emph{arXiv preprint arXiv:1810.04805}.

\bibitem[{Eisenstein and Barzilay(2008)}]{eisenstein2008bayesian}
Jacob Eisenstein and Regina Barzilay. 2008.
\newblock Bayesian unsupervised topic segmentation.
\newblock In \emph{Proceedings of the 2008 Conference on Empirical Methods in
  Natural Language Processing}, pages 334--343.

\bibitem[{El-Kishky et~al.(2014)El-Kishky, Song, Wang, Voss, and
  Han}]{el2014scalable}
Ahmed El-Kishky, Yanglei Song, Chi Wang, Clare~R Voss, and Jiawei Han. 2014.
\newblock Scalable topical phrase mining from text corpora.
\newblock \emph{Proceedings of the VLDB Endowment}, 8(3):305--316.

\bibitem[{Hochreiter and Schmidhuber(1997)}]{hochreiter1997long}
Sepp Hochreiter and J{\"u}rgen Schmidhuber. 1997.
\newblock Long short-term memory.
\newblock \emph{Neural computation}, 9(8):1735--1780.

\bibitem[{Hong et~al.(2015{\natexlab{a}})Hong, Wang, Ortiz, and
  Whitehouse}]{hong2015building}
Dezhi Hong, Hongning Wang, Jorge Ortiz, and Kamin Whitehouse.
  2015{\natexlab{a}}.
\newblock The building adapter: Towards quickly applying building analytics at
  scale.
\newblock In \emph{BuildSys}.

\bibitem[{Hong et~al.(2015{\natexlab{b}})Hong, Wang, and
  Whitehouse}]{hong2015clustering}
Dezhi Hong, Hongning Wang, and Kamin Whitehouse. 2015{\natexlab{b}}.
\newblock Clustering-based active learning on sensor type classification in
  buildings.
\newblock In \emph{Proceedings of the 24th ACM International on Conference on
  Information and Knowledge Management}, pages 363--372. ACM.

\bibitem[{Jiao et~al.(2020)Jiao, Li, Wu, Hong, Gupta, and
  Shang}]{jiao2020senser}
Yang Jiao, Jiacheng Li, Jiaman Wu, Dezhi Hong, Rajesh Gupta, and Jingbo Shang.
  2020.
\newblock Senser: Learning cross-building sensor metadata tagger.
\newblock In \emph{Proceedings of the 2020 Conference on Empirical Methods in
  Natural Language Processing: Findings}, pages 950--960.

\bibitem[{Karpathy et~al.(2015)Karpathy, Johnson, and
  Fei-Fei}]{karpathy2015visualizing}
Andrej Karpathy, Justin Johnson, and Li~Fei-Fei. 2015.
\newblock Visualizing and understanding recurrent networks.
\newblock \emph{arXiv preprint arXiv:1506.02078}.

\bibitem[{Kim et~al.(2016)Kim, Jernite, Sontag, and Rush}]{kim2016character}
Yoon Kim, Yacine Jernite, David Sontag, and Alexander~M Rush. 2016.
\newblock Character-aware neural language models.
\newblock In \emph{Thirtieth AAAI Conference on Artificial Intelligence}.

\bibitem[{Kiros et~al.(2014)Kiros, Salakhutdinov, and
  Zemel}]{kiros2014multimodal}
Ryan Kiros, Ruslan Salakhutdinov, and Rich Zemel. 2014.
\newblock Multimodal neural language models.
\newblock In \emph{International Conference on Machine Learning}, pages
  595--603.

\bibitem[{Koc et~al.(2014)Koc, Akinci, and Berg{\'e}s}]{koc2014comparison}
Merthan Koc, Burcu Akinci, and Mario Berg{\'e}s. 2014.
\newblock Comparison of linear correlation and a statistical dependency measure
  for inferring spatial relation of temperature sensors in buildings.
\newblock In \emph{BuildSys}, pages 152--155. ACM.

\bibitem[{Koh et~al.(2018)Koh, Balaji, Sengupta, McAuley, Gupta, and
  Agarwal}]{koh2018scrabble}
Jason Koh, Bharathan Balaji, Dhiman Sengupta, Julian McAuley, Rajesh Gupta, and
  Yuvraj Agarwal. 2018.
\newblock Scrabble: transferrable semi-automated semantic metadata
  normalization using intermediate representation.
\newblock In \emph{Proceedings of the 5th Conference on Systems for Built
  Environments}, pages 11--20. ACM.

\bibitem[{Manning et~al.(2014)Manning, Surdeanu, Bauer, Finkel, Bethard, and
  McClosky}]{manning2014stanford}
Christopher~D Manning, Mihai Surdeanu, John Bauer, Jenny~Rose Finkel, Steven
  Bethard, and David McClosky. 2014.
\newblock The stanford corenlp natural language processing toolkit.
\newblock In \emph{Proceedings of 52nd annual meeting of the association for
  computational linguistics: system demonstrations}, pages 55--60.

\bibitem[{Opitz and Burst(2019)}]{opitz2019macro}
Juri Opitz and Sebastian Burst. 2019.
\newblock Macro f1 and macro f1.
\newblock \emph{arXiv preprint arXiv:1911.03347}.

\bibitem[{Parameswaran et~al.(2010)Parameswaran, Garcia-Molina, and
  Rajaraman}]{parameswaran2010towards}
Aditya Parameswaran, Hector Garcia-Molina, and Anand Rajaraman. 2010.
\newblock Towards the web of concepts: Extracting concepts from large datasets.
\newblock \emph{Proceedings of the VLDB Endowment}, 3(1-2):566--577.

\bibitem[{Peters et~al.(2018)Peters, Neumann, Iyyer, Gardner, Clark, Lee, and
  Zettlemoyer}]{peters2018deep}
Matthew~E Peters, Mark Neumann, Mohit Iyyer, Matt Gardner, Christopher Clark,
  Kenton Lee, and Luke Zettlemoyer. 2018.
\newblock Deep contextualized word representations.
\newblock \emph{arXiv preprint arXiv:1802.05365}.

\bibitem[{Pritoni et~al.(2015)Pritoni, Bhattacharya, Culler, and
  Modera}]{pritoni2015short}
Marco Pritoni, Arka~A Bhattacharya, David Culler, and Mark Modera. 2015.
\newblock Short paper: A method for discovering functional relationships
  between air handling units and variable-air-volume boxes from sensor data.
\newblock In \emph{BuildSys}, pages 133--136. ACM.

\bibitem[{Qi et~al.(2018)Qi, Dozat, Zhang, and
  Manning}]{qi-etal-2018-universal}
Peng Qi, Timothy Dozat, Yuhao Zhang, and Christopher~D. Manning. 2018.
\newblock \href {https://doi.org/10.18653/v1/K18-2016} {Universal dependency
  parsing from scratch}.
\newblock In \emph{Proceedings of the {C}o{NLL} 2018 Shared Task: Multilingual
  Parsing from Raw Text to Universal Dependencies}, pages 160--170, Brussels,
  Belgium. Association for Computational Linguistics.

\bibitem[{Qi et~al.(2020)Qi, Zhang, Zhang, Bolton, and Manning}]{qi2020stanza}
Peng Qi, Yuhao Zhang, Yuhui Zhang, Jason Bolton, and Christopher~D Manning.
  2020.
\newblock Stanza: A python natural language processing toolkit for many human
  languages.
\newblock \emph{arXiv preprint arXiv:2003.07082}.

\bibitem[{Schumann et~al.(2014)Schumann, Ploennigs, and
  Gorman}]{schumann2014towards}
Anika Schumann, Joern Ploennigs, and Bernard Gorman. 2014.
\newblock Towards automating the deployment of energy saving approaches in
  buildings.
\newblock In \emph{Proceedings of the 1st ACM Conference on Embedded Systems
  for Energy-Efficient Buildings}, pages 164--167. ACM.

\bibitem[{Sch{\"u}tze et~al.(2008)Sch{\"u}tze, Manning, and
  Raghavan}]{schutze2008introduction}
Hinrich Sch{\"u}tze, Christopher~D Manning, and Prabhakar Raghavan. 2008.
\newblock Introduction to information retrieval.
\newblock In \emph{Proceedings of the international communication of
  association for computing machinery conference}, volume~4.

\bibitem[{Settles(2009)}]{settles2009active}
Burr Settles. 2009.
\newblock Active learning literature survey.
\newblock Technical report, University of Wisconsin-Madison Department of
  Computer Sciences.

\bibitem[{Shang et~al.(2018)Shang, Liu, Gu, Ren, Ren, and
  Han}]{shang2018learning}
Jingbo Shang, Liyuan Liu, Xiaotao Gu, Xiang Ren, Teng Ren, and Jiawei Han.
  2018.
\newblock Learning named entity tagger using domain-specific dictionary.
\newblock In \emph{Proceedings of the 2018 Conference on Empirical Methods in
  Natural Language Processing}, pages 2054--2064.

\bibitem[{Shi et~al.(2019)Shi, Newsham, Chen, and Gunay}]{shi2019evaluation}
Zixiao Shi, Guy~R Newsham, Long Chen, and H~Burak Gunay. 2019.
\newblock Evaluation of clustering and time series features for point type
  inference in smart building retrofit.
\newblock In \emph{Proceedings of the 6th ACM International Conference on
  Systems for Energy-Efficient Buildings, Cities, and Transportation}, pages
  111--120.

\bibitem[{Weng and Agarwal(2012)}]{weng2012buildings}
Thomas Weng and Yuvraj Agarwal. 2012.
\newblock From buildings to smart buildings—sensing and actuation to improve
  energy efficiency.
\newblock \emph{IEEE Design \& Test of Computers}, 29(4):36--44.

\end{thebibliography}
